%% file: main.tex
\documentclass[conference]{IEEEtran}
\IEEEoverridecommandlockouts
\usepackage{cite}
\usepackage{amsmath,amssymb,amsfonts}
\usepackage{algorithmic}
\usepackage{graphicx}
\usepackage{textcomp}
\usepackage{xcolor}

\usepackage{multirow}
\usepackage{wrapfig}
\usepackage{booktabs}       
\usepackage{url}            

\def\BibTeX{{\rm B\kern-.05em{\sc i\kern-.025em b}\kern-.08em
    T\kern-.1667em\lower.7ex\hbox{E}\kern-.125emX}}
\begin{document}

\title{SMPL Normal Map Is All You Need for Single-view Textured Human Reconstruction}

\author{\IEEEauthorblockN{Wenhao Shen\textsuperscript{1*}, Gangjian Zhang\textsuperscript{2*}, Jianfeng Zhang\textsuperscript{3}, Yu Feng\textsuperscript{2}, Nanjie Yao\textsuperscript{4}, Xuanmeng Zhang\textsuperscript{5}, Hao Wang\textsuperscript{1\dag}\thanks{\textsuperscript{*}Equal contribution.}\thanks{\textsuperscript{\dag}Corresponding author.}}
\IEEEauthorblockA{
\textsuperscript{1}\textit{Nanyang Technological University}, Singapore \\
\textsuperscript{2}\textit{The Hong Kong University of Science and Technology (Guangzhou)}, China \\
\textsuperscript{3}\textit{National University of Singapore}, Singapore 
 \textsuperscript{4}\textit{Zhejiang University of Technology}, China  \\
 \textsuperscript{5}\textit{University of Technology Sydney}, Australia \\
Email: wenhao005@ntu.edu.sg, gzhang292@connect.hkust-gz.edu.cn}
}

\maketitle

\begin{abstract}
Single-view textured human reconstruction aims to reconstruct a clothed 3D digital human by inputting a monocular 2D image. Existing approaches include feed-forward methods, limited by scarce 3D human data, and diffusion-based methods, prone to erroneous 2D hallucinations. To address these issues, we propose a novel SMPL normal map Equipped 3D Human Reconstruction (SEHR) framework, integrating a pretrained large 3D reconstruction model with human geometry prior. SEHR performs single-view human reconstruction without using a preset diffusion model in one forward propagation. Concretely, SEHR consists of two key components: SMPL Normal Map Guidance (SNMG) and SMPL Normal Map Constraint (SNMC). SNMG incorporates SMPL normal maps into an auxiliary network to provide improved body shape guidance. SNMC enhances invisible body parts by constraining the model to predict an extra SMPL normal Gaussians. Extensive experiments on two benchmark datasets demonstrate that SEHR outperforms existing state-of-the-art methods.
\end{abstract}


\input{sec/1_introduction}

\input{sec/2_related_work}
\input{sec/3_method}

\input{sec/4_experiments}
\input{sec/5_conclusion}
 


\bibliographystyle{IEEEbib}
\bibliography{main}

\end{document}

%% file: sec/1_introduction.tex
\section{Introduction}
\label{intro}

With the increasing popularity of virtual worlds, there is a growing need to create realistic digital humans, which play a role in AR/VR technology. However, creating a 3D clothed avatar from a monocular image has been an ill-posed problem and a persistent challenge since a single view can hardly provide sufficient information. Strong ambiguity exists in the geometry and texture of the human body's invisible parts.

Previous methods for this task can be categorized into two branches. The first branch~\cite{xiu2022icon, xiu2023econ, zhang2024global_gta, Zhang_2024_sifu} involves training a single-forward human reconstruction model based on limited 3D human scans, combining prior human body knowledge, such as SMPL~\cite{loper2015smpl} to enhance the reconstruction quality. 
However, due to the scarcity of 3D human body data~\cite{tao2021function4d_thuman}, these methods can only train a small-scale network and lack sufficient generalizability. 
The second branch, exemplified by SiTH~\cite{ho2024sith} and TeCH~\cite{huang2024tech}, utilizes techniques like ControlNet~\cite{zhang2023adding} and Score Distillation Sampling~\cite{poole2022dreamfusion} to exploit the knowledge of 2D Stable Diffusion models~\cite{Rombach_2022_CVPR_sd15} for 3D generation. However, the lack of 3D knowledge in these Stable Diffusion models may lead to many issues in the generated 3D human body, including bad hallucination and Janus (multi-faced) problem.

Meanwhile, data-driven approaches are increasingly emerging in the generic 3D reconstruction task. They often necessitate large 3D datasets for training a diffusion model capable of generating novel views from a monocular input and training a single view or multi-view large 3D object reconstruction model~\cite{tang2024lgm, hong2023lrm}. For example, LGM~\cite{tang2024lgm} uses a pre-trained novel view synthesis model to produce object multi-view RGB images, followed by a U-Net~\cite{ronneberger2015u_unet} trained from massive 3D objects to reconstruct 3D Gaussians~\cite{3DGaussian}.

However, directly applying the aforementioned 3D large models to human scope is non-trivial. Given the small-scale available 3D human datasets~\cite{tao2021function4d_thuman}, training large human synthesis and reconstruction models is hard. To resolve this, we propose a novel \textbf{S}MPL normal map \textbf{E}quipped 3D \textbf{H}uman \textbf{R}econstruction (\textbf{SEHR}) framework. SEHR is designed and finetuned based on the pre-trained large object Gaussians reconstruction model, LGM~\cite{tang2024lgm}, which is rich in 3D knowledge. 
Meanwhile, to avoid the possible Janus problem incurred by the multi-view Stable Diffusion model, SEHR only receives one front image as input. As compensation, we combine the monocular SMPL estimation technique~\cite{pavlakos2019expressive_smplx, cai2023smplerx} with the LGM for providing the necessary human depth/structure information lacking in the single-view RGB input. 

Notably, SEHR is distinct from previously prevailing single-forward human reconstruction methods~\cite{xiu2022icon, xiu2023econ, zhang2024global_gta, Zhang_2024_sifu}. SEHR can be integrated with off-the-shelf large 3D reconstruction models to provide effective human shape guidance, hereby facilitating the adaptability of these powerful pretrained large object models for human reconstruction.

Concretely, as shown in Fig.~\ref{ours_pipeline}, we inject human geometry prior in the form of orthogonal SMPL normal maps. We design a parallel network SNMG to guide learning accurate 3D positions of the human Gaussians in the main branch. To improve human invisible back areas, we also propose a new objective SNMC featuring reconstructing extra SMPL normal Gaussians. Experiments on two out-of-distribution test sets validate the effectiveness of the proposed solution.

Our contributions can be summarized as follows: 
\begin{itemize}
    \item We explore the adaptability of large 3D reconstruction models for single-view human reconstruction and propose a new framework, SEHR, without any preset novel-view Stable Diffusion model.
    \item We propose a new approach, SNMG, which integrates the human prior with a Gaussian reconstruction model by incorporating SMPL normal maps into the auxiliary network to improve the geometry of human Gaussians.
    \item We propose a new training objective, SNMC, which enhances the network's ability to learn human invisible parts by constraining it to predict SMPL normal Gaussians.
    \item SEHR outperforms existing state-of-the-art methods on benchmark datasets.
\end{itemize}

%% file: sec/2_related_work.tex
 \begin{figure*} 
	\centering
	\includegraphics[width=0.95\linewidth]{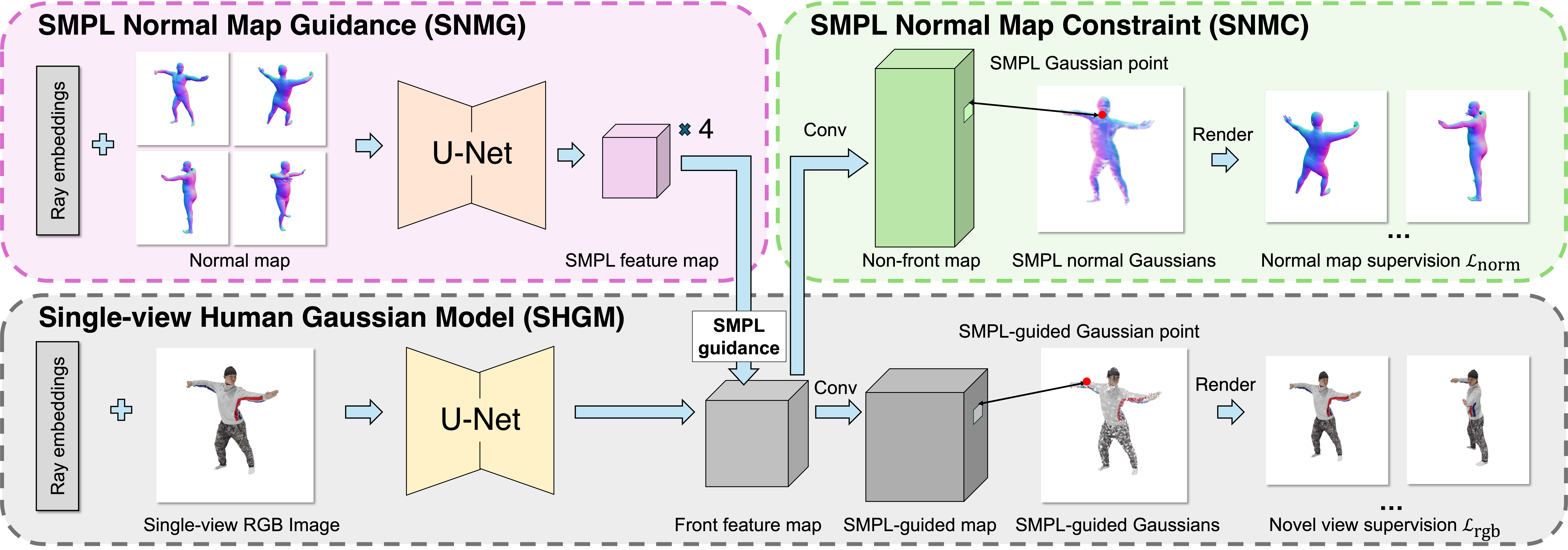}
 	\vspace{-0.2cm}
	\caption{\textbf{Overview of SEHR.} Based on the SHGM model, SEHR introduces an extra branch to extract the human prior in the orthogonal SMPL normal maps and provide shape guidance for Gaussian learning, called SMPL Normal Map Guidance (SNMG). Additionally, we propose using the back, left, and right views of SMPL normal maps to constrain the hallucination of invisible body parts, named SMPL Normal Map Constraint (SNMC). }
	\label{ours_pipeline}
	\vspace{-0.3cm} 
\end{figure*}

\section{Related Work}
\label{related_work}

\subsection{Single-View 3D Human Reconstruction}
The earliest method of single-view 3D human reconstruction, PIFu~\cite{saito2019pifu} and its extension PIFu-HD \cite{saito2020pifuhd} introduce a pixel-aligned implicit function that enables shape and texture generation. 
ARCH~\cite{huang2020arch} employs a rigged canonical space to facilitate the re-posing. 
ICON \cite{xiu2022icon} introduces the skinned body model \cite{loper2015smpl}, while ECON \cite{xiu2023econ} and 2K2K \cite{han2023high_2k2k} regard the point cloud as the geometry prior to obtain the mesh. 
PHORHUM\cite{alldieck2022photorealistic} enhances the geometry modeling by combining the estimation of scene illumination and surface albedo. 
The latest SiTH \cite{ho2024sith} addresses information scarcity in unseen regions by using an SD model. 
However, these methods are constrained by limited data availability and the lack of a robust backbone capable of providing comprehensive 3D knowledge for human generation, resulting in suboptimal outcomes.

\subsection{Large 3D Object Reconstruction Model} 
Recent popular solutions for 3D object reconstruction involve large object reconstruction models, which adopt convolution~\cite{ronneberger2015u_unet} or transformer networks as the backbone, using their capacity to achieve 2D image to 3D object reconstruction in a feed-forward way. The breaking work of LRM~\cite{hong2023lrm}, increasing both model capacity and data volume, enables direct reconstruction from one image. Subsequent works like~\cite{xu2023dmv3d} combine with the multi-view SD model to achieve better results. Meanwhile, many works~\cite{hong2023lrm, tang2024lgm} are trying to find a better 3D representation, such as 3D Gaussians~\cite{3DGaussian}. 
Our work aims to apply the flourishing of 3D object reconstruction to the human field by incorporating human-prior knowledge into the large object reconstruction model in response to the scarcity of 3D human data.

%% file: sec/3_method.tex
\section{Method}
\label{method}

\subsection{Preliminaries}
\label{prel}
\textbf{Gaussian splatting.} 3D Gaussians splatting~\cite{3DGaussian} is a collection of 3D Gaussians, denoted by $\Theta$, to represent 3D data. 
Each 3D Gaussian is defined by a set of parameters to indicate its characteristics: $\Theta_i=\left\{\mathbf{x}_i, \mathbf{s}_i, \mathbf{q}_i, \alpha_i, \mathbf{c}_i\right\} \in \mathbb{R}^{14}$, 
in which $\mathbf{x} \in \mathbb{R}^3$, $\mathbf{s} \in \mathbb{R}^3$, $\mathbf{q} \in \mathbb{R}^4$, $\alpha \in \mathbb{R}$ and $\mathbf{c} \in \mathbb{R}^C$ denote the center, scaling factor, rotation quaternion, opacity value and color feature respectively. In our work, $C$ is set to $3$ for rendering, where spherical harmonics can be used to model view-dependent effects. These Gaussians can be rendered using the differentiable rasterizer.

\textbf{Large multi-view Gaussian model (LGM).} Our work extends the LGM~\cite{tang2024lgm}, a large-scale reconstruction model for general objects using 3D Gaussian Splatting~\cite{3DGaussian}. LGM takes four orthogonal RGB images with camera ray embeddings as input and employs a U-Net to produce four unique feature maps, finally combining them into 3D Gaussians. These Gaussians are rendered from novel viewpoints and compared to ground truth images for supervision. In our approach, we adapt the original LGM framework by limiting the input to a single view, making it specifically suited for the task of single-view 3D human reconstruction.

\subsection{Single-View Human Gaussian Model (SHGM)}~\label{SHGM}

Note that LGM~\cite{tang2024lgm} first relies on an off-the-shelf multi-view diffusion model to produce multi-view images. 
However, existing multi-view diffusion models struggle with creating multi-view consistent images, introducing artifacts in subsequent 3D generation phases.
To this end, we modify the input from four views to a single view and propose a Single-view Human Gaussian Model (SHGM) adapting LGM to our single-view human reconstruction task.

Concretely, the input of SHGM is a feature map $\textbf{F}$ obtained from the concatenation of the front view RGB image $\textbf{I}_{f}$ and its Pl\"{u}cker ray embedding $\textbf{R}_{f}$. 
Ray embedding encodes the camera poses and is parameterized as the cross-product of the camera ray origin and direction vectors.
We further feed $\textbf{F}$ into the asymmetric U-Net to obtain a low-resolution feature map $\textbf{F}_{l}$ whose each pixel corresponds to one Gaussian. Finally, $\textbf{F}_{l}$ goes through a convolution layer to produce output feature map $\textbf{F}_{o}$, which is used as the Gaussians parameter set $\Theta$. 
To supervise the generated Gaussians and reduce the computation cost, we use a differentiable renderer \cite{3DGaussian} to render them. We render the RGB and mask images of eight views, including one input view and seven novel views, from $\Theta$. We compute MSE and LPIPS loss on rendered RGB images and MSE loss on masked images:
\begin{equation} \label{equa1}
\mathcal{L}_\mathrm{rgb}=\mathcal{L}_{\mathrm{mse}}\left(I_{\mathrm{rgb}}, I_{\mathrm{rgb}}^{\mathrm{GT}}; I_{\mathrm{mask}}, I_{\mathrm{mask}}^{\mathrm{GT}}\right)+\mathcal{L}_{\mathrm{lpips}}\left(I_{\mathrm{rgb}},I_{\mathrm{rgb}}^{\mathrm{GT}}\right).
\end{equation}

\subsection{SMPL Normal Map Guidance (SNMG)}
\label{SNMG}
SHGM serves as our baseline model. With only monocular input, optimizing SHGM becomes more challenging compared to the original LGM. To mitigate this, we introduce SNMG as an enhancement branch, injecting essential human prior—specifically, SMPL estimated from monocular images. This provides spatial structural information about the human body, guiding SHGM towards improved Gaussian representations for subsequent 3D generation.

\textbf{SMPL normal input.} We first leverage an off-the-shelf human mesh recovery method SMPLer-X~\cite{cai2023smplerx} to estimate SMPL mesh from the input RGB image.
Then we render four orthogonal SMPL normal maps from the estimated SMPL mesh.
SNMG takes these normal maps concatenated with their corresponding ray embeddings (front, back, left, and right) as input. 
It is notable that the front-view SMPL normal map shares the same camera pose as the input RGB image.
To reduce the memory cost, we set each input SMPL normal map size as half of the input image size. Then, these rendered normal maps are fed into an extra U-Net and obtain $\textbf{F}_{n} = \{\textbf{F}_{nf}, \textbf{F}_{nb}, \textbf{F}_{nl}, \textbf{F}_{nr}\}$. 

\textbf{SMPL-guided feature maps.} 
Since SHGM, which is our main reconstruction branch, only contains the color information, we involve $\textbf{F}_{n}$ into SHGM to provide human body shape guidance.
Technically, we adopt the residual connection strategy. We first use the bilinear approach to up-sample $\textbf{F}_{n}$ to the same resolution as the RGB feature map $\textbf{F}_{l}$. Then, we residually add $\textbf{F}_{n}$ with $\textbf{F}_{l}$ and adopt a convolution layer to obtain four SMPL-guided feature maps $\textbf{F}_{g}=\{\textbf{F}_{gf}, \textbf{F}_{gb}, \textbf{F}_{gl}, \textbf{F}_{gr}\}$. 

\textbf{SMPL-guided Gaussians.} Based on work~\cite{szymanowicz2023splatter}, the large reconstruction model achieves the mapping from 2D RGB image to 3D Gaussians. So, we parallelly conduct the SMPL normal map mapping into the same space, where we integrate the color feature with the SMPL shape feature to obtain Gaussians containing structure knowledge. These improved points can fit accurate 3D spatial positions. We obtain these improved 3D Gaussians by collecting the SMPL-guided feature maps from four views since each of them contains only the partial shape-guided color Gaussian. Four of them are fused to obtain the high-resolution output Gaussian feature map $\textbf{F}_{o}$, with each pixel containing parameters of one final Gaussian.

\subsection{SMPL Normal Map Constraint (SNMC)}
\label{SNMC}

Although SNMG effectively guides the Gaussian points, we find that the generated 3D human shapes still exhibit irregularities and random hollows in invisible body parts. 
We already have RGB and normal maps for the front-view texture and geometry optimization, while the other three views only contain normal information, leading to imbalanced optimization. 
So, SNMC aims to enhance the reconstruction quality of non-front invisible views by using the non-front SMPL normal maps (back, left, and right) as additional supervision signals.

\textbf{SMPL Gaussian points.} 
We predict three non-front SMPL Gaussian feature maps through another convolution using the residual connection of the non-front parts of $\textbf{F}_{n}$ and $\textbf{F}_{l}$.
The non-front feature maps are reshaped to create the SMPL normal Gaussian set ${\Theta}^{'}$. Notably, ${\Theta}^{'}$ differs from ${\Theta}$ of SHGM by replacing the color feature with the direction feature.

\textbf{SMPL Gaussian supervision.} Based on SMPL normal Gaussians~${\Theta}^{'}$, we use another differentiable renderer to render three non-front SMPL normal maps $I_{\mathrm{n}}$. We supervise them with corresponding ground-truth SMPL normal maps $I_{\mathrm{n}}^{\mathrm{GT}}$: 
\begin{equation} \label{equa2}
\mathcal{L_{\mathrm{norm}}}=\mathcal{L}_{\mathrm{mse}}\left(I_{\mathrm{n}}, I_{\mathrm{n}}^{\mathrm{GT}}\right)+\mathcal{L}_{\mathrm{lpips}}\left(I_{\mathrm{n}}, I_{\mathrm{n}}^{\mathrm{GT}}\right).
\end{equation}

Finally, our training objective is:
\begin{equation} \label{equa3}
\mathcal{L_{\mathrm{f}}}=\mathcal{L}_\mathrm{rgb}+\mathcal{L_{\mathrm{norm}}}.
\end{equation}

%% file: sec/4_experiments.tex
\begin{table}
\centering
	\caption{\textbf{Comparison with SOTAs on CustomHumans and CAPE.} \\
    ``$^{\ast}$" denotes the models trained on extra commercial data, and ``$^{\dagger}$" denotes the model trained on the THuman2.1 dataset while others use THuman2.0. }
	\label{tab:custom}
 			\vspace{-0.20cm}
 \scalebox{0.9}
	{
  		\begin{tabular}{l|ccc|c}
			\toprule
			\multirow{4}*{Method} &  \multicolumn{4}{|c}{ CustomHumans~\cite{ho2023customhuman} } \\
   \cmidrule{2-5}
			&  \begin{tabular}{c} 
CD: P-to-S / \\
S-to-P $(\mathrm{cm}) \downarrow$
\end{tabular} & NC $\uparrow$ & f-Score $\uparrow$ & \begin{tabular}{c} 
LPIPS: F \\
$\left(\times 10^{-2}\right) \downarrow$
\end{tabular} \\
			\midrule
PIFu$^{\ast}$~\cite{saito2019pifu} & $2.209 / 2.582$ & $0.805$ & $34.881$ & $6.073 / 8.496$ \\
PIFuHD$^{\ast}$~\cite{saito2020pifuhd}& $2.107 / 2.228$ & $0.804$ & $39.076$ & -  \\
PaMIR$^{\ast}$~\cite{zheng2020pamir} & $2.181 / 2.507$ & $0.813$ & $35.847$ & $4.646 / 
7.152$ \\
2K2K$^{\ast}$~\cite{han2023high_2k2k} & $2.488 / 3.292$ & $0.796$ & $30.186$ & -  \\
FOF$^{\ast}$~\cite{li2022neurips_fof} & $2.079 / 2.644$ & $0.808$ & $36.013$ & - \\
			\midrule

ICON~\cite{xiu2022icon} & $2.256 / 2.795$ & $0.791$ & $30.437$ & - \\
ECON~\cite{xiu2023econ} & $2.483 / 2.680$ & $0.797$ & $30.894$ & - \\
GTA~\cite{zhang2024global_gta} & $2.349 / 2.568$ & $0.794$ & $31.028$ & $6.234 / 7.692$  \\
SIFU~\cite{Zhang_2024_sifu} & $2.332 / 2.546$ & $0. 798$ & $30.146$ & $5 . 9 43 / 7.649$  \\
SiTH~\cite{ho2024sith} & $1 . 8 7 1 / 2 . 0 4 5$ & $0. 8 2 6$ & $37.029$ & $3 . 9 2 9 / 6 . 8 0 3$   \\
LGM~\cite{tang2024lgm} & $2.085 / 2.851$ & $0.794$ & $30.389$ & $5.383/7.599$ \\
			\midrule

\underline{Ours} & $1.262 / 1.666$ & 0.848 & 47.978 & 3.710/6.052 \\
\underline{Ours$^{\dagger}$} & $ \textbf{1.206} / \textbf{1.588}$ & $\textbf{0.853}$ & $\textbf{50.008}$ & $\textbf{3.259}/\textbf{5.534}$ \\
			\midrule
   			\multirow{1}*{ } &  \multicolumn{4}{|c}{ CAPE~\cite{ma2020cape} } \\
			\midrule
PIFu$^{\ast}$~\cite{saito2019pifu} & $2.368 / 3.763$ & $0.778$ & $33.842$ & $2.720$ \\
PIFuHD$^{\ast}$~\cite{saito2020pifuhd}  & $2.401 / 3.522$ & $0.772$ & $35.706$ & -  \\
PaMIR$^{\ast}$~\cite{zheng2020pamir}  & $2.190 / 2.806$ & $0.804$ & $36.725$ & $2.085$ \\
2K2K$^{\ast}$~\cite{han2023high_2k2k} & $2.478 / 3.683$ & $0.782$ & $28.700$ & - \\
FOF$^{\ast}$~\cite{li2022neurips_fof} & $2.196 / 4.040$ & $0.777$ & $34.227$ & - \\
			\midrule
ICON~\cite{xiu2022icon}  & $2.516 / 3.079$ & $0.786$ & $29.630$ & - \\
ECON~\cite{xiu2023econ} & $2.475 / 2.970$ & $0.788$ & $30.488$ & - \\
GTA~\cite{zhang2024global_gta} & $2.635 / 2.965$ & $0.784$&$29.296$& $3.022$\\
SIFU~\cite{Zhang_2024_sifu} & $2.619 / 2.998$ & $0.786$&$29.020$& $2.930$ \\
SiTH~\cite{ho2024sith} & $1.899 / 2.261$ & $0.816$&$37.763$&$1.977$  \\
LGM~\cite{tang2024lgm} & $2.587 / 4.372$ & $0.759$ & $27.709$ & $2.041$  \\
			\midrule
\underline{Ours} & $1.764 / \textbf{1.783}$ & $0.820$ & $39.052$ & $\textbf{1.933}$ \\
\underline{Ours$^{\dagger}$} & $\textbf{1.734} / 1.845$ & $\textbf{0.826}$ & $\textbf{41.866}$ & $2.033$\\
			\bottomrule
  \end{tabular}
	}
			\vspace{-0.20cm}
\end{table}

 \section{Experiments}
\label{Experiments}

\subsection{Implementation Details}
\label{imple}

\textbf{Training.} We adopt the training setup from LGM~\cite{tang2024lgm}. The pretrained LGM model is fine-tuned on 3D human data using a single NVIDIA A100 GPU. For each batch, we sample four orthogonal views and four random views to generate 8-view RGB images at a resolution of $512 \times 512$. One of the four orthogonal views is selected as the input view. Using the same camera settings, we render four orthogonal SMPL normal maps from the ground truth SMPL mesh. The AdamW optimizer is used with a learning rate of ${10}^{-4}$.

\textbf{Datasets.} To ensure a fair comparison with previous methods, we adopt the dataset settings used in SiTH~\cite{ho2024sith}. By default, our method is trained on the THuman2.0~\cite{tao2021function4d_thuman} dataset and evaluated on out-of-distribution results using the CAPE~\cite{ma2020cape} and CustomHumans~\cite{ho2023customhuman} datasets.

\textbf{Inference.} Only the front view human color image is required for inference. Moreover, four orthogonal SMPL normal maps will be rendered from the SMPL mesh estimated by SMPLer-X~\cite{cai2023smplerx}. Note that our method outputs 3D Gaussians, which can be further transformed into meshes~\cite{tang2024lgm} for compatibility with existing pipes.

\textbf{Evaluation metrics.} We follow previous work~\cite{xiu2022icon, ho2024sith} regarding evaluation metrics and protocols. For the 3D metrics, we utilize Chamfer Distance (CD) (both predictions to scans and vice versa), Normal Consistency (NC), and f-Score on the generated meshes. Besides, we employ the LPIPS metric to compare the front and back texture rendering of the generated meshes with the GT mesh rendering.

\subsection{Quantitative Comparison}

\textbf{Table~\ref{tab:custom} shows competitive performance of SEHR against the SOTA methods on CAPE~\cite{ma2020cape} and CustomHumans~\cite{ho2023customhuman} benchmarks.} 
We reproduce the results of GTA~\cite{zhang2024global_gta} and SIFU~\cite{Zhang_2024_sifu} since they do not open their code.
On CustomHumans, our method surpasses the SOTA by margins of $0.609$cm and $0.379$cm at CD P2S and S2P, with increases in NC and f-Score of $0.022$ and $10.949$. The improvements in LPIPS for front and back renderings are $0.219*10^{-2}$ and $0.751*10^{-2}$, respectively. Our method also outperforms SOTAs on CAPE, with improvements in CD P2S and S2P of $0.135$cm and $0.478$cm, and increases in f-Score of $1.289$. Training on the public THuman2.1 dataset could lead to further improvements. 

\textbf{Table~\ref{tab:time} shows the efficiency of our method.} SEHR has a much faster inference speed than previous methods.

\begin{table}
\centering
	\caption{\textbf{Inference time comparison.} }
	\label{tab:time}
 			\vspace{-0.2cm}
 \scalebox{0.90}
	{
		\begin{tabular}{l|ccccc}
			\toprule
   Method & ECON~\cite{xiu2023econ} & GTA~\cite{zhang2024global_gta} & SIFU~\cite{Zhang_2024_sifu} & SiTH~\cite{ho2024sith}  & Ours \\
			\midrule
   Duration & 8min30s & 4min20s & 1min5s & 1min25s  & \textbf{35s} \\
			\bottomrule
		\end{tabular}
	}
 \vspace{-0.2cm}
\end{table}

\subsection{Ablation Study} 
\textbf{Table~\ref{tab:snmg} proves the effectiveness of SNMG.} It is evident that SNMG outperforms SHGM, which was trained using a single-view RGB image, in all metrics on both CAPE and CustomHumans. Moreover, incremental improvement is observed with an increase in the number of SMPL normal maps used. The rationale is as the number of SMPL normal maps increases, the effect of SNMG in providing SMPL guidance for human Gaussian optimization becomes more pronounced. Therefore, the final reconstruction quality also becomes better.

\textbf{Table~\ref{tab:snmc} validates the effectiveness of SNMC.} 
It is found that ``Three-view SNMC" is most effective compared to ``Three-view HNNC" and ``Four-view SNMC", showing superiority on both datasets. The rationale is that the single-view reconstruction is an ill-posed problem, which means that relying solely on a front view for full body reconstruction lacks necessary information, such as the human back shape. However, ``Three-view HNNC" forces the model to predict or hallucinate the invisible shape without supporting information, potentially leading to incorrect fitting of the current human body data distribution and not performing well on other datasets. This explains the slight drop in performance on the other dataset trained with THuman2.0.
It is also observed that the ``Three-view SNMC" outperforms the ``Four-view SNMC". The reason is the SNMC is designed to enhance learning in parts with insufficient information, such as the back view, in contrast to the front view, which contains more information (both RGB image and SMPL normal map) and is easier to learn. In the case of the ``Four-view" approach, the additional front view for front view shape constraint may compromise the learning of the other three views, resulting in the SNMC not achieving the desired performance. As indicated in Table~\ref{tab:snmc}, the incorporation of front view constraint leads to slight performance drops.

\begin{table}
\centering
	\caption{\textbf{Ablation of SNMG.} Compared to the full SNMG with all ``four-view" SMPL (front, back, left, right) normal maps to provide guidance, in the “Front-view only” method, we only use the Front-view as guidance.}
 
	\label{tab:snmg}

 	\scalebox{0.81}
	{
		\begin{tabular}{l|ccc}
			\toprule

   			\multirow{2}*{Method} &  \multicolumn{3}{|c}{ CustomHumans } \\
      \cmidrule{2-4}
			&  \begin{tabular}{c} 
CD: P2S /
S2P $(\mathrm{cm}) \downarrow$
\end{tabular} & NC $\uparrow$ & f-Score $\uparrow$  \\
			\midrule

SHGM & $2.085 / 2.851$ & 0.794 & 30.389  \\
\quad \quad \quad + SNMG (Front-view only) & $1.713 / 2.258$ & 0.817 & 36.076  \\
\quad \quad \quad + SNMG (Four-view) & $\textbf{1.290 / 1.700}$ & \textbf{0.846} & \textbf{46.095} \\
			\midrule
   \midrule

			\multirow{2}*{Method} &  \multicolumn{3}{|c}{ CAPE } \\
   \cmidrule{2-4}
			&  \begin{tabular}{c} 
CD: P2S / S2P $(\mathrm{cm}) \downarrow$
\end{tabular} & NC $\uparrow$ & f-Score $\uparrow$  \\
			\midrule
SHGM & $2.587 / 4.372$ & 0.759 & 27.709  \\
\quad \quad \quad + SNMG (Front-view only) & $2.280 / 3.229$ & 0.780 & 32.525 \\
\quad \quad \quad + SNMG (Four-view) & $\textbf{1.898 / 2.008}$ & \textbf{0.804} & \textbf{37.351}  \\

			\bottomrule
		\end{tabular}
	}
			\vspace{-0.2cm}

\end{table}

\begin{table}
\centering
	\caption{\textbf{Ablation of SNMC.} ``HNMC(Human Normal Map Constraint)" means the model is supervised by human normal maps. ``SNMC(SMPL Normal Map Constraint)" is supervised by SMPL normal maps. ``Three-view" uses non-front views(back, left, right) for supervision. ``four-view" uses all four views.}
	\label{tab:snmc}

	\scalebox{0.75}{
		\begin{tabular}{l|ccc}
			\toprule
			\multirow{2}*{Method} &  \multicolumn{3}{|c}{ CustomHumans } \\
   \cmidrule{2-4}
			&  \begin{tabular}{c} 
CD: P2S / S2P $(\mathrm{cm}) \downarrow$
\end{tabular} & NC $\uparrow$ & f-Score $\uparrow$  \\
			\midrule
SHGM + SNMG & $1.290 / 1.700$ & 0.846 & 46.095  \\
\quad \quad \quad \quad \quad \quad \quad + Three-view HNMC & $1.300 / 1.714$ & 0.847 & 46.054  \\
\quad \quad \quad \quad \quad \quad \quad + Four-view SNMC & $1.309 / 1.702$ & 0.845 & 46.530  \\
\quad \quad \quad \quad \quad \quad \quad + Three-view SNMC & $\textbf{1.262 / 1.666}$ & \textbf{0.848} & \textbf{47.978}  \\
			\midrule
   \midrule

			\multirow{2}*{Method} &  \multicolumn{3}{|c}{ CAPE } \\
   \cmidrule{2-4}
			&  \begin{tabular}{c} 
CD: P2S / S2P $(\mathrm{cm}) \downarrow$
\end{tabular} & NC $\uparrow$ & f-Score $\uparrow$  \\
			\midrule

SHGM  + SNMG & $1.898 / 2.008$ & 0.804 & 37.351  \\
\quad \quad \quad \quad \quad \quad \quad + Three-view HNMC & $1.784 / 1.807$ & 0.817 & 38.916  \\
\quad \quad \quad \quad \quad \quad \quad + Four-view SNMC & $1.779 / 1.807$ & 0.817 & 38.965 \\
\quad \quad \quad \quad \quad \quad \quad+ Three-view SNMC & $\textbf{1.764 / 1.783}$ & \textbf{0.820} & \textbf{39.052}  \\
			\bottomrule
		\end{tabular}
	}
	\vspace{-0.4cm}

\end{table}

\begin{figure*} 
	\centering
	\includegraphics[width=0.85\linewidth]{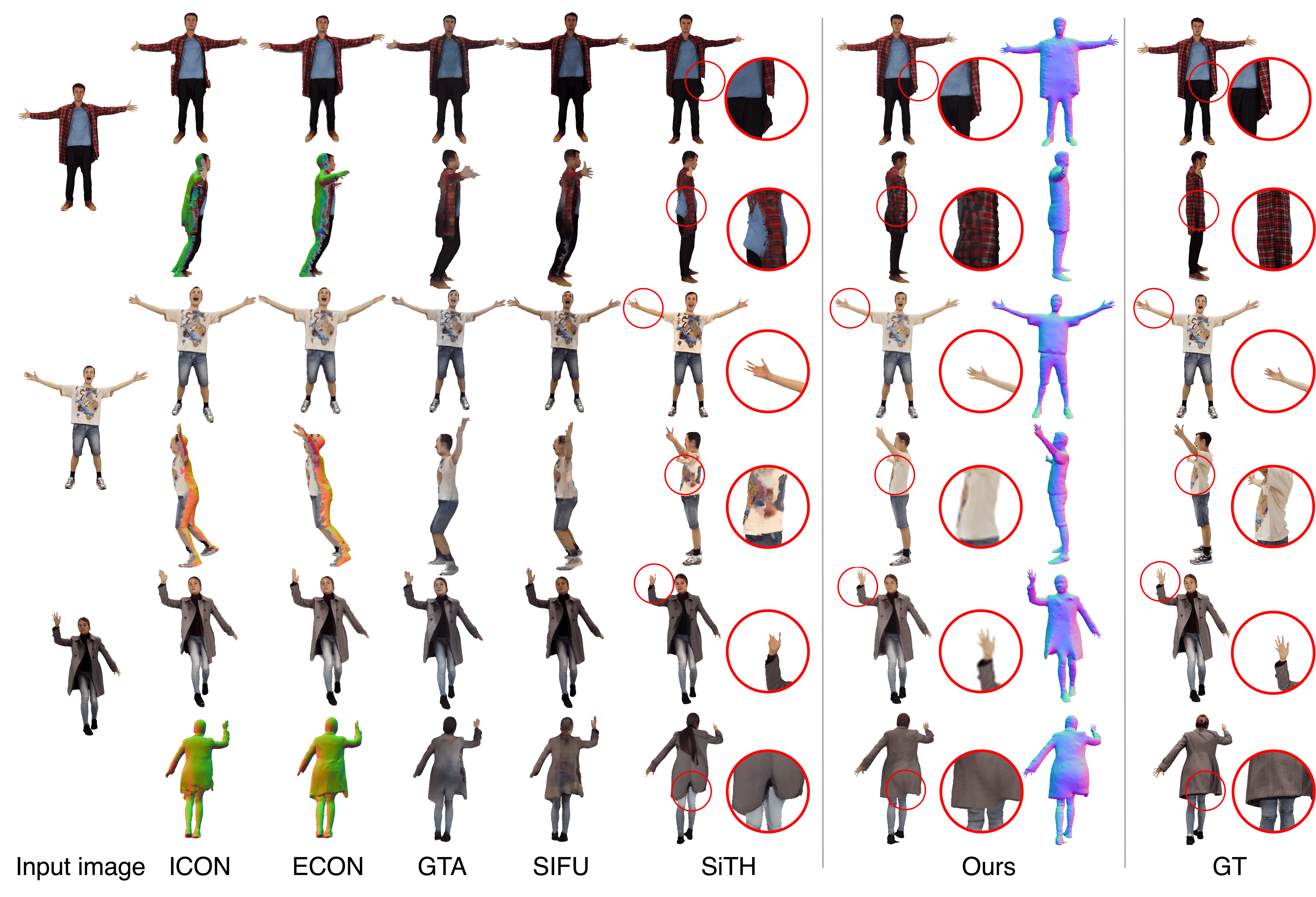}
	\vspace{-0.2cm}
	\caption{\textbf{Visualization comparison with SOTA methods.} It is shown that our method has a better reconstruction effect on the details of the 3D human body, such as the fingers and edges of the clothes. Meanwhile, our method also has fewer hallucination artifacts compared to SiTH~\cite{ho2024sith}, which wrongly generates patterns that do not exist on clothes. Please zoom in for a better view.}
	\label{sotavis}
\end{figure*}

 \begin{figure} 
  		\vspace{-0.4cm}

	\centering
	\includegraphics[width=1.0\linewidth]{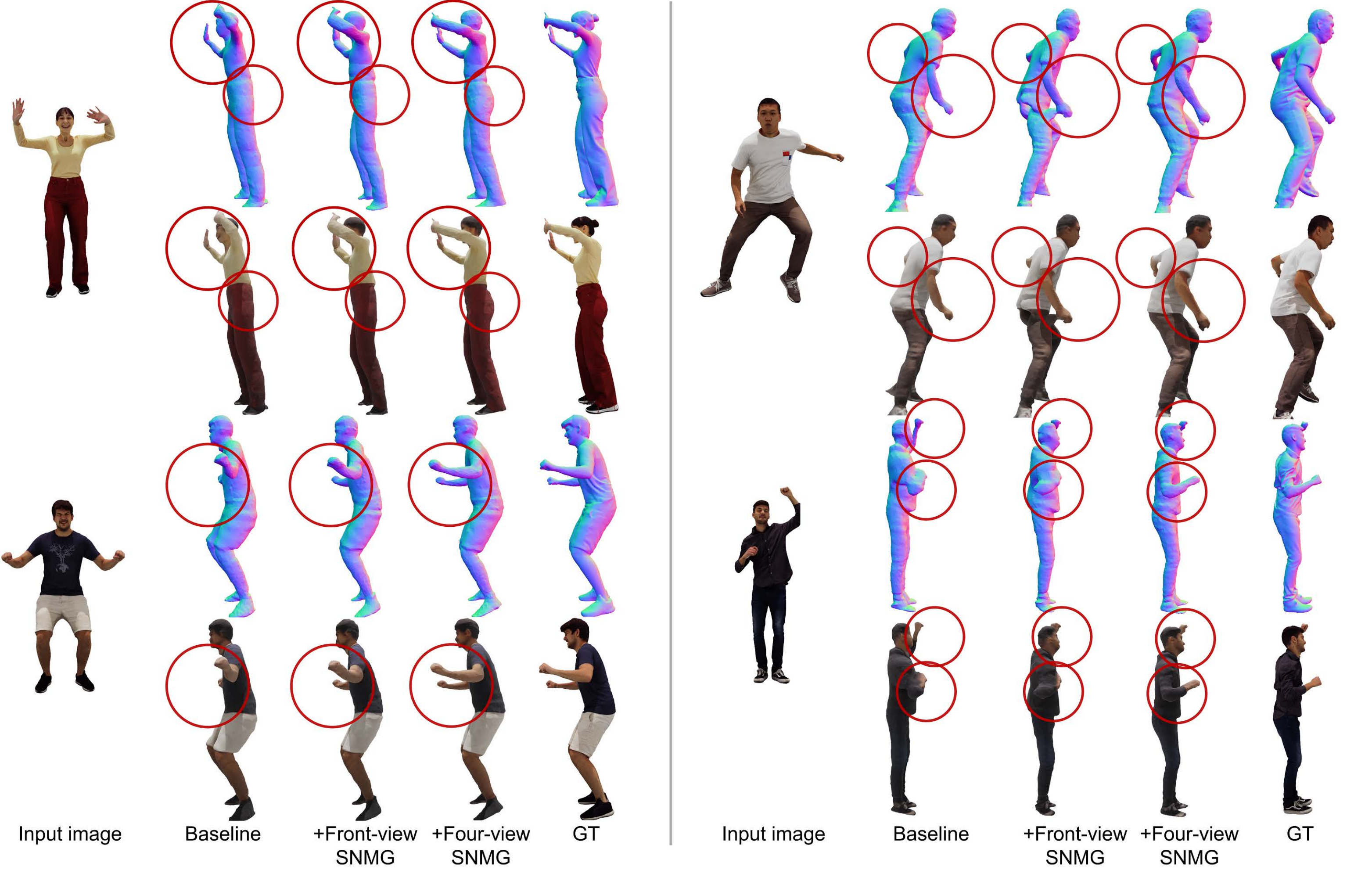}
 	\vspace{-0.6cm}
	\caption{\textbf{Visual ablation of SNMG.} As the number of views increases, SNMG guidance becomes increasingly effective in refining the final 3D Gaussian/mesh. For instance, the stretched-out human hand better aligns with the labeled shape.}
	\label{vissnmg}
		\vspace{-0.4cm}
\end{figure}

 \begin{figure} 
   		\vspace{-0.4cm}
	\centering
	\includegraphics[width=1.0\linewidth]{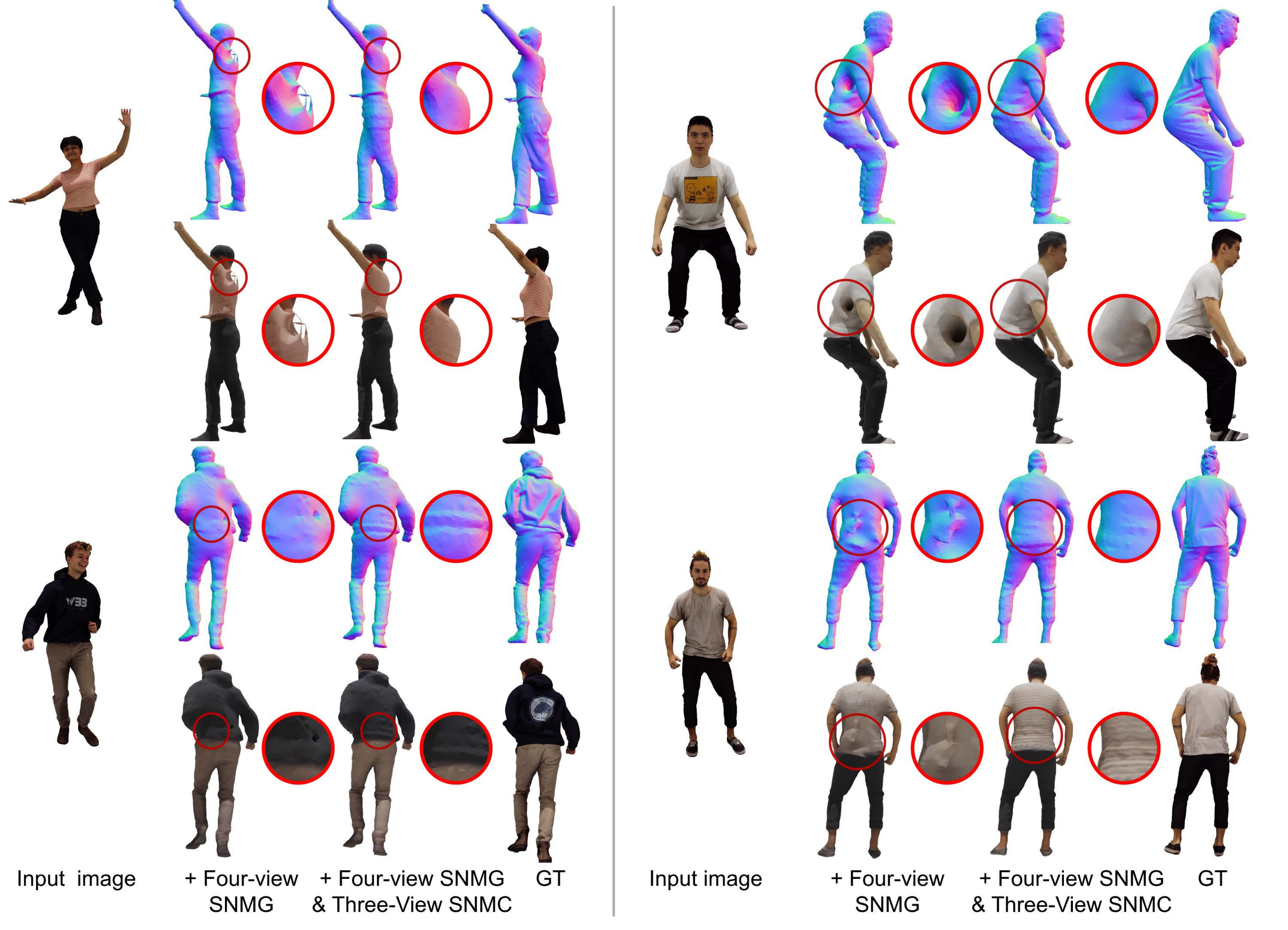}
        \vspace{-0.6cm}
	\caption{\textbf{Visual ablation of SNMC.} We compare the methods with and without SNMC. While SNMG significantly enhances the overall shape of the reconstructed 3D mesh, certain subtle details, such as the shoulder hull in the first case, remain slightly inaccurate. In contrast, SNMC further improves both shape and texture quality, better aligned with the ground truth. }
	\label{vissnmc}
		\vspace{-0.4cm}
\end{figure}

\subsection{Qualitative Results} 

\textbf{Fig.~\ref{sotavis} visualizes our advantages over SOTA methods.} 
We can observe significant improvement in human geometry, particularly in the finger parts. This is achieved through our approach, which effectively integrates prior knowledge of the human body into large-scale object reconstruction models, leveraging their robust reconstruction capabilities while enhancing the precision of human-specific details. 
Furthermore, our method generates fewer hallucinations compared to Stable Diffusion-based models such as SiTH\cite{ho2024sith}. The single-view feed-forward approach we employ avoids reliance on preset Stable Diffusion models, thereby mitigating errors that could compromise the quality of 3D reconstruction.

\textbf{Fig.~\ref{vissnmg} illustrates the quality enhancement brought by SNMG.}
The input image remains unchanged across comparisons, emphasizing the differences in human shape between various methods and the ground truth. As the number of views in SNMG increases, the reconstructed human shape progressively aligns with the labeled true mesh. These results confirm the effectiveness of SNMG in guiding accurate human shape reconstruction. Furthermore, they highlight the previously underappreciated importance of SMPL normal maps in human reconstruction. Effectively utilizing these normal maps can lead to substantial improvements in reconstruction quality.

\textbf{Fig.~\ref{vissnmc} demonstrates the effectiveness of SNMC.} Keeping the input image constant, we compare human shape and texture across different methods against the ground truth. Notably, occluded details, such as shoulder and back contours, are inadequately captured using only the SNMG setting. To overcome this limitation, we introduced SNMC. The visual comparisons clearly illustrate that SNMC effectively resolves these issues, significantly enhancing both shape and texture reconstruction quality.

%% file: sec/5_conclusion.tex
\section{Conclusion}
\label{conclusion}

This paper presents SEHR, a novel framework developed from existing large object models to address the challenging task of single-view textured human reconstruction. 
The framework comprises two essential components: SNMG and SNMC. SNMG utilizes human SMPL normal maps within an auxiliary network to provide precise shape guidance for Gaussian-based human reconstruction. 
Meanwhile, SNMC applies constraints to the network, enabling the reconstruction of occluded human regions through SMPL Gaussian points. Experimental results on two public datasets highlight the superior performance of our framework.

\section{Acknowledgment}
This research is supported by the RIE2025 Industry Alignment Fund – Industry Collaboration Projects (IAF-ICP) (Award I2301E0026), administered by A*STAR, as well as supported by Alibaba Group and NTU Singapore through Alibaba-NTU Global e-Sustainability CorpLab (ANGEL).